# FEATHer: Fourier-Efficient Adaptive Temporal Hierarchy Forecaster for Time-Series Forecasting

Jaehoon Lee†, Seungwoo Lee†, Younghwi Kim†, Dohee Kim*, *Member*, Sunghyun Sim*, *Member*

**Abstract**— Time-series forecasting plays a fundamental role in industrial domains such as manufacturing, energy management, logistics, and smart factory operations. As these systems increasingly evolve toward cyber-physical automation, forecasting models are required to operate directly on edge devices, including programmable logic controllers, embedded microcontrollers, and other low-power industrial hardware. These environments impose strict constraints on latency, memory, and energy consumption, limiting deployable models to at most a few thousand parameters. Under such constraints, conventional deep forecasting architectures become computationally impractical. To address these challenges, we propose the Fourier-Efficient Adaptive Temporal Hierarchy Forecaster (FEATHer), a multiscale temporal model designed to achieve accurate long-term forecasting under severe resource limitations. FEATHer introduces four key components: (i) an ultra-lightweight multiscale temporal decomposition that transforms the input sequence into point-, high-, mid- and low-frequency pathways, (ii) a shared Dense Temporal Kernel that performs efficient temporal mixing via projection–depthwise convolution–projection without recurrence or attention, (iii) a frequency-aware branch gating mechanism that adaptively fuses multiscale representations based on the spectral characteristics of the normalized input sequence and (iv) a Sparse Period Kernel that reconstructs long-horizon outputs through period-wise downsampling and shared linear transformation to capture periodic and seasonal structures with minimal additional parameters. Compared to existing baselines, FEATHer maintains an exceptionally compact architecture while achieving strong predictive performance and operates in extreme parameter-budget regimes, including settings with as few as 400 trainable parameters. Across eight long-term forecasting benchmarks and multiple horizons, FEATHer achieves the best overall ranking, recording 60 first-place results with an average rank of 2.05. These results demonstrate that reliable long-range forecasting is achievable even under highly constrained edge conditions and suggest a practical direction for next-generation industrial systems requiring real-time inference with minimal computational cost.

**Index Terms**— Time-series Forecasting, Edge AI, Ultra-Lightweight Models, Fourier-Efficient Adaptive Temporal Hierarchy Forecaster (FEATHer)

──────────◆──────────

## 1 INTRODUCTION

Time-series forecasting is a core capability in modern industrial intelligence [1], enabling production scheduling [2], anomaly detection [3], predictive maintenance [4], energy balancing [5], safety monitoring [6], and process control across manufacturing [7], logistics [8], transportation infrastructure [9]. Accurate long-horizon forecasting is increasingly critical as industrial environments transition toward cyber-physical and autonomous operation, where decisions must be made continuously using streaming sensor data [10].

At the same time, practical deployment increasingly requires forecasting models to run directly on edge platforms, such as programmable logic controllers (PLCs), embedded microcontrollers, industrial IoT sensors, and low-power gateways, under strict constraints on latency, memory, and energy [11]. These devices typically offer limited CPU throughput and small memory footprints while still demanding millisecond-level response times and reliable behavior under nonstationary operating conditions [12]. Under such constraints, large Transformer-based architectures, deep convolutional encoders, and even moderately sized state-space models are often impractical for real-world deployment [13]. This motivates the need for resource-constrained forecasting models that maintain accuracy over long horizons under extremely tight parameter budgets [14].

Recent research has explored lightweight forecasting architectures, including DLinear [15], TiDE [16], TSMixer [17], FITS [18], CycleNet [19], and SparseTSF [20]. While these methods demonstrate the potential of linear decomposition, shallow temporal mixing, and period-aware sparse projections, several challenges remain when

This work was supported by the National Research Foundation of Korea (NRF) grant funded by the Korean government (MSIT) (No.RS-2023-00218913) and was supported by Basic Science Research Program through the National Research Foundation of Korea(NRF) funded by the Ministry of Education(No.RS-2025-25396743) (Co-first author: Jeahoon Lee, Seungwoo Lee, and Younghwi Kim, Co-corresponding author: Sunghyun Sim and Dohee Kim). The source code is publicly available at: https://github.com/WDSLab/FEATHer

Jeahoon Lee, Seongwoo Lee, and Younghwi Kim are with the Graduate School of AI Convergence Engineering, Changwon National University, 20 Changwondaehak-ro, Ui-chang Gu, 51140 Changwon, South Korea (e-mail: dynamic97312@naver.com, and dudgnl6032@naver.com

Dohee Kim and Sunghyun Sim are with the Departments of AI Convergence, Changwon National University, 20 Changwondaehak-ro, Ui-chang Gu, 51140 Changwon, South Korea (e-mail: ssh@changwon.ac.kr and kimdohee@changwon.ac.kr

applying them to industrial long-horizon forecasting. First, many designs rely on a single temporal scale or a fixed periodic structure, which is insufficient to represent the hierarchical patterns commonly observed in industrial signals, ranging from rapid fluctuations to medium-range transitions and long-term seasonal drifts [21]. Second, lightweight architecture often lacks explicit mechanism for structured frequency decomposition, forcing heterogeneous temporal components into a single representational pathway [22]. This can induce cross-frequency interference and degrade temporal resolution [23]. Third, despite being labeled lightweight, many models still require tens of thousands of parameters, which may exceed the extreme budgets imposed by tightly constrained edge controllers [24].

To address these challenges, we propose Fourier-Efficient Adaptive Temporal Hierarchy Forecaster (FEATHer), an ultra-lightweight model designed for accurate long-horizon forecasting under severe resource constraints. FEATHer follows a structured design principle in which representations are explicitly organized across temporal scales and adaptively fused based on the spectral characteristics of the input. Specifically, FEATHer generates multiscale representations using lightweight temporal filtering via depthwise operations to obtain point-, high-, mid-, and low-frequency pathways. Each pathway is processed by a shared Dense Temporal Kernel (DTK), composed of a linear projection, a depthwise temporal convolution, and a reverse projection, enabling efficient temporal mixing without recurrence or self-attention. FEATHer further incorporates a frequency-aware gating module that dynamically reweights multiscale pathways by analyzing the spectrum of the normalized input, improving robustness under nonstationary dynamics. For long-horizon forecasting, FEATHer employs a Sparse Period Kernel (SPK) that reconstructs periodic and seasonal structure through phase-aligned and period-aligned reorganization with a shared linear transformation, enabling effective long-horizon modeling with minimal additional parameters.

Owing to these architectural choices, FEATHer operates under an ultra-compact parameter budget, for example, fewer than 1,000 parameters in compact configurations, while delivering competitive accuracy across standard long-horizon forecasting benchmarks. These results suggest that FEATHer is practical for real-time industrial edge deployment under stringent latency, memory, and energy constraints. The key contributions of this work are as follows:

- **(C1)** We introduce an ultra-lightweight multiscale decomposition that separates input dynamics into frequency-aligned pathways, enabling hierarchical temporal modeling under extreme parameter budgets while reducing cross-frequency interference.
- **(C2)** We propose the DTK, a shared lightweight mixing block that captures temporal dependencies without recurrence or self-attention, preventing parameter growth in multi-branch designs.
- **(C3)** We develop a frequency-aware gating mechanism that adaptively fuses multiscale representations based on the input spectrum, enabling structured adaptation to nonstationary dynamics without additional learnable parameters.
- **(C4)** We design the SPK for phase-aligned and period-aligned long-horizon reconstruction, capturing periodic and seasonal structure without increasing model depth.
- **(C5)** We show that FEATHer achieves competitive performance on long-horizon forecasting benchmarks while maintaining an ultra-compact footprint, supporting practical deployment in constrained industrial edge environments.

The remainder of this paper is organized as follows. Section 2 reviews related work on lightweight forecasting. Section 3 describes the FEATHer architecture and its components. Section 4 presents theoretical analyses of stability, expressiveness, and computational efficiency. Section 5 reports the experimental setup and results. Section 6 presents ablation studies and empirical validation of the theoretical analysis. Section 7 concludes with limitations and future directions.

## 2 RELATED WORKS

Long-term time series forecasting (LTSF) has been studied from multiple perspectives, including **(i)** architectures designed to capture long-range dependencies, **(ii)** efficiency-oriented models that reduce parameter and computational budgets, and **(iii)** methods that introduce explicit temporal structure, such as multiscale decomposition, spectral modeling, or periodic reconstruction. This section reviews representative approaches and highlights the remaining gap addressed by FEATHer, especially in the Sub-1K parameter regime required by severely resource-constrained industrial edge devices.

### 2.1 Modeling Long-Horizon Dependencies

A substantial body of LTSF research focuses on capturing long-range temporal dependencies using Transformer-based mechanisms [25]. Transformer-based LTSF models, including Autoformer [26], FEDformer [27], PatchTST [28], and iTransformer [29], aim to improve long-horizon forecasting by enhancing the modeling of long-range dependencies through attention-based mechanisms [30]. These approaches incorporate decomposition-driven autocorrelation, frequency-domain attention, patch-wise representations, or instance-token reformulations to increase expressive capacity over long sequences [30].

Although these models achieve strong benchmark performance, they rely on attention operations and deep architectural stacks that typically require parameter budgets ranging from hundreds of thousands to millions [31]. As a result, practical deployment in memory- and latency-constrained environments such as PLCs, microcontrollers, and industrial edge devices remains challenging [32].

## 2.2 Efficiency-Oriented Forecasting Models

To reduce model complexity, a parallel line of research explores efficiency-oriented forecasting architectures that aim to achieve competitive accuracy with reduced parameter and computational budgets. Linear and decomposition-based methods, such as DLinear and NLinear, demonstrate that strong performance can be achieved through well-structured linear mappings without reliance on attention mechanisms [15]. Beyond purely linear models, compact architectures with shallow mixing layers or MLP-based components, including TiDE [16] and TSMixer [17], seek to enhance expressive power while maintaining efficiency.

Frequency-domain and cycle-aware approaches offer an additional efficiency-oriented direction. Methods such as FITS leverage spectral filtering and interpolation to support long-range forecasting with lower computational cost than full attention-based decoders [18], while lightweight cycle-aware designs such as CycleNet exploit repeating temporal structures to improve extrapolation without deep decoding stacks [19]. Despite these advances, an important limitation remains: many models described as lightweight still operate with parameter budgets on the order of several thousand to tens of thousands, which can be prohibitive for severely resource-constrained edge platforms [34]. Moreover, because most efficiency-oriented architectures process temporal dynamics through a single scale or pathway, they can suffer from representational bottlenecks when mixed-frequency patterns such as rapid fluctuations, mid-range transitions, and slow drifts must be modeled simultaneously under extremely tight parameter budgets.

## 2.3 Structured Multi-Scale and Spectral Modeling

A complementary line of research introduces explicit temporal structure, such as multiscale decomposition, spectral modeling, or periodic reconstruction, to improve long-horizon stability and interpretability [33]-[35]. Multi-resolution convolutional designs, wavelet-inspired architectures, and hierarchical models seek to disentangle temporal patterns through structured transformations [36]. Related approaches incorporate periodic or sparse cross-period mappings to mitigate parameter growth as forecast horizons increase, as exemplified by methods such as SparseTSF, which leverage period-aware sparse projections to model periodic components efficiently [20]. Despite these advances, existing structured models exhibit several limitations when considered for edge-grade deployment under ultra-compact model budgets. First, many approaches rely on fixed or coarse decomposition rules that do not explicitly enforce a clean separation of temporal components into distinct frequency bands, leaving residual cross-frequency interference unresolved. Second, adaptive mechanisms that dynamically modulate the relative importance of temporal scales are less commonly explored in ultra-compact settings; when present, they often operate in the time domain, for example via pooled statistics, rather than directly exploiting spectral signatures that more naturally reflect frequency composition. Third, models primarily designed for periodic reconstruction may perform well on seasonal signals but often lack robustness for nonstationary and mixed-frequency sequences in which periodicity coexists with abrupt changes and drifting trends.

## 2.4 Positioning of FEATHer

FEATHer is motivated by a fundamental tension between forecasting accuracy and deployability in LTSF. High-capacity LTSF models can capture long-range dependencies with high fidelity, but their reliance on deep architectures and attention mechanisms renders them impractical for deployment in resource-constrained industrial edge environments. Conversely, lightweight models are deployable on edge hardware but often lack structured mechanisms to handle mixed-frequency and nonstationary dynamics under Sub-1K parameter budgets. We argue that this gap arises from three key unresolved challenges.

- **(R1) Explicit band-wise separation under extreme parameter budgets:** Most lightweight forecasting architectures process temporal dynamics through a single pathway or rely on coarse, fixed decompositions, making it difficult to isolate heterogeneous components such as instantaneous fluctuations, high-frequency oscillations, mid-range motifs, and low-frequency drifts without residual cross-frequency interference.
- **(R2) Adaptive scale selection grounded in spectral characteristics:** Even when multiscale representations are available, ultra-compact models rarely incorporate principled mechanisms that adaptively reweigh temporal scales based on the spectral composition of the input. As a result, robustness degrades when dominant dynamics vary across instances or operating regimes.
- **(R3) Parameter-optimal reconstruction of long-horizon periodic structure**: Many existing approaches allocate horizon-dependent parameters or rely on iterative decoding strategies, both of which are undesirable for edge deployment. Efficient long-horizon modeling instead requires horizon-agnostic structures that capture periodic and seasonal patterns with minimal additional parameters.

To address these challenges, FEATHer adopts a unified design that integrates **(i)** a structured multiscale temporal decomposition that produces band-specialized pathways, **(ii)** a shared DTK that enables efficient temporal mixing across all bands without recurrence or attention, **(iii)** a frequency-aware gating mechanism that adaptively fuses multiscale representations based on spectral signatures, and **(iv)** a SPK that reconstructs long-range periodic components through compact, period-aligned mappings. Together, these components enable accurate and robust long-horizon forecasting while remaining compatible with the severe latency, memory, and energy constraints of industrial edge hardware.

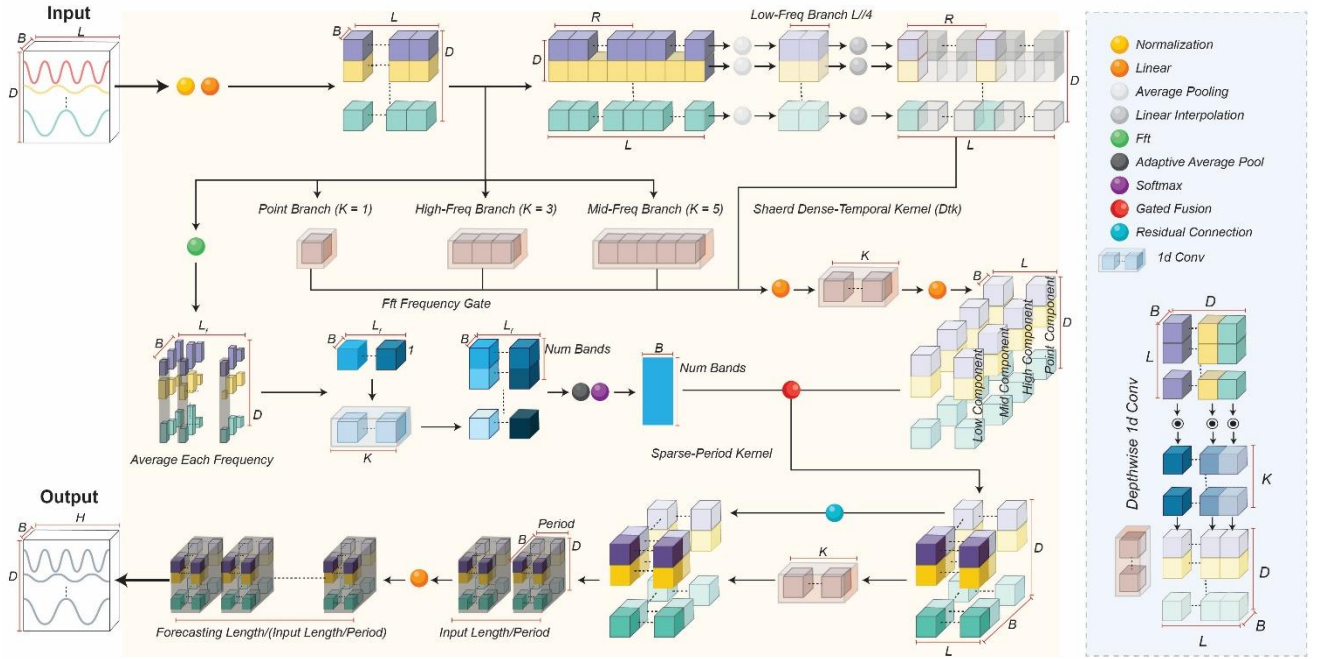

**Fig. 1.** Overall architecture of FEATHer

## 3 FOURIER-EFFICIENT ADAPTIVE TEMPORAL HIERARCHY FORECASTER

This section presents the overall architecture of the Fourier-Efficient Adaptive Temporal Hierarchy Forecaster (FEATHer). FEATHer is designed to support stable long-horizon forecasting under stringent latency, memory, and energy constraints typical of industrial edge devices. The model integrates four core components: (i) a *structured multiscale temporal decomposition* that produces complementary scale representations, (ii) a shared *DTK* that performs efficient temporal mixing through linear projections and depthwise temporal filtering, (iii) a *frequency-aware branch gating* module that adaptively fuses scale representations based on the input spectrum and (iv) a *SPK* that reconstructs long-horizon periodic and seasonal structure through compact period-aligned transformations. FEATHer avoids computationally intensive recurrence and self-attention, relying instead on lightweight linear operators, depthwise filtering, and sparse period-wise mappings, making it suitable for deployment on resource-constrained hardware. Fig. 1 illustrates the overall architecture of FEATHer.

### 3.1 Problem Setup and Notation

We consider multivariate forecasting with an input window of length $L$, a forecasting horizon $H$, and $D$ variables. The input sequence is denoted by $X = [x_1, \ldots x_L]^T \in \mathbb{R}^{L \times D}$, and the target horizon is $Y = [y_1, \ldots y_H]^T \in \mathbb{R}^{H \times D}$. FEATHer produces $\hat{Y} \in \mathbb{R}^{H \times D}$ from X. The model constructs $B \in \{2,3,4\}$ scale pathways; the active branch set is denoted by $\mathcal{B} \subseteq \{p, h, m, l\}$, corresponding to point, high, mid, and low frequency pathways, where $|\mathcal{B}| = B$. The DTK uses a latent width $S$, and the SPK uses a period $P$. All branch representations are aligned to the same temporal length $L$ so that they can be fused without cross-scale alignment overhead. In this work, scale refers to frequency-aligned temporal representations obtained via lightweight temporal filtering.

### 3.2 Structured Multiscale Temporal Decomposition

FEATHer begins by transforming X into multiple scale representations that emphasize different temporal behaviors while remaining computationally inexpensive.

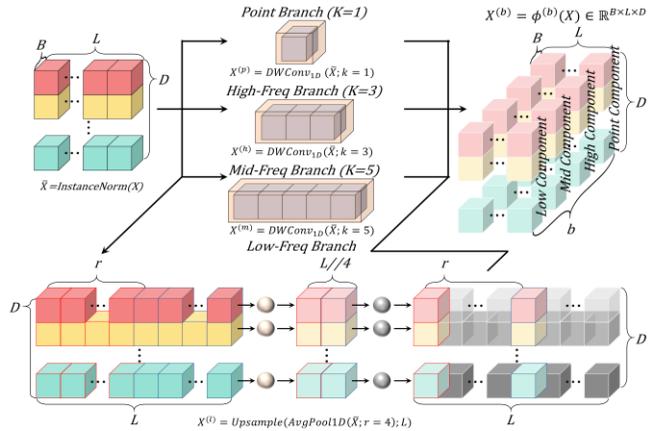

**Fig. 2.** Structure of Multiscale Temporal Decomposition

For each active branch $b \in \mathcal{B}$, we generate $X^{(b)} = \phi^{(b)}(X) \in \mathbb{R}^{L \times D}$. where $\phi^{(b)}(\cdot)$ is implemented using lightweight depthwise temporal operations. The point branch preserves instantaneous information through a kernel-size-1 depthwise operator, while the high and mid branches use short-support depthwise convolutions, for example, kernel sizes 3 and 5, to emphasize local and medium-range variations, respectively. The low branch isolates slow components by temporally downsampling the input using average pooling with stride *r* and then interpolating back to length *L*, yielding an explicit low-pass bias without introducing heavy parameters. This decomposition provides complementary pathways that reduce cross-frequency interference under compact model budgets while keeping all representations time-aligned for subsequent fusion.

### 3.3 DTK

Each decomposed pathway is processed by a shared DTK that performs efficient temporal mixing without recurrence or self-attention. DTK adopts a projection–depthwise filtering–inverse projection structure. For a branch input $X^{(b)} \in \mathbb{R}^{L \times D}$, DTK first projects the input to a compact latent width $S$

$$Z^{(b)} = X^{(b)} W_{in} \in \mathbb{R}^{L \times S} \quad (1)$$

where $W_{in} \in \mathbb{R}^{D \times S}$. A depthwise temporal convolution then mixes information along the temporal dimension independently for each latent channel, where $k_t$ denotes the depthwise temporal convolution kernel.

$$U^{(b)} = \text{DWConv}_t(Z^{(b)}; k_t) \in \mathbb{R}^{L \times S} \quad (2)$$

$$H^{(b)} = U^{(b)} W_{out} \in \mathbb{R}^{L \times D} \quad (3)$$

The same DTK parameters $\{W_{in}, W_{out}, DWConv\ weights\}$ are shared across all branches $b \in \mathcal{B}$. This parameter sharing prevents growth in model size as the number of branches increases and keeps the architecture ultra-lightweight, while still allowing each branch to express scale-specific temporal dynamics through its distinct input signal $X^{(b)}$

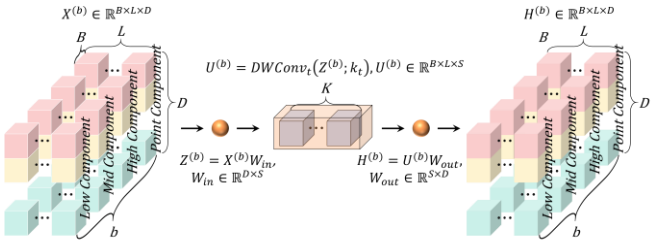

**Fig. 3** Structure of Dense Temporal Kernel

### 3.4 Frequency-aware Branch Gating

The importance of each temporal scale varies across instances and operating regimes. FEATHer therefore employs a frequency-aware gating mechanism that assigns branch weights based on the input spectral profile, enabling instance-adaptive fusion while remaining lightweight. We first normalize the input sequence along the temporal dimension for each variable. We then compute a real FFT along the temporal axis,

$$F = \mathcal{F}(\bar{X}), \quad (4)$$

and take the magnitude spectrum

$$A = |F| \quad (5)$$

A compact spectral descriptor is obtained by averaging across variables,

$$a = \frac{1}{D} \sum_{d=1}^{D} A_{:,d} \in \mathbb{R}^{L_f} \quad (6)$$

where $L_f = \lfloor L/2 \rfloor + 1$. A lightweight gating network $\psi(\cdot)$ maps a to branch logits,

$$u = \psi(a) \in \mathbb{R}^{B} \quad (7)$$

and a softmax operation yields branch weights,

$$g = \text{softmax}(u) \in \mathbb{R}^{B} \quad (8)$$

The fused representation is formed by weighting the DTK outputs:

$$H = \sum_{b \in \mathcal{B}} g_b H^{(b)} \in \mathbb{R}^{L \times D} \quad (9)$$

In practice, $\psi(\cdot)$ can be implemented as a small Conv1D stack operating on a. Because the descriptor length is only $L_f$, this module introduces negligible overhead while allowing the model to emphasize the most informative scales for each input.

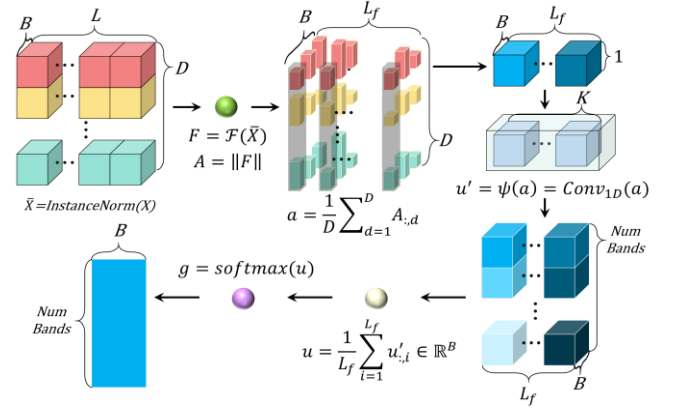

**Fig. 4** Structure of Frequency-aware Branch Gating

### 3.5 SPK for Long-Horizon Reconstruction

To produce $\hat{Y}$ efficiently, FEATHer employs a SPK that reconstructs periodic and seasonal structure through compact period-aligned transformations. Starting from $H \in \mathbb{R}^{L \times D}$, SPK first applies a depthwise temporal aggregation:

$$H'(t) = (H * w)(t) \quad (10)$$

where w is a learnable 1D filter with padding that preserves the temporal length *L*. SPK then reorganizes $H'$ into phase-aligned groups using a period *P*. When *L* and *H* are divisible by *P*, we define $n=L/P$ and $m=H/P$. For each variable *d*, $H'_{:,d}$ is reshaped into $\widetilde{H}_d \in \mathbb{R}^{P \times n}$, where each row corresponds to a fixed phase within the period. A shared linear mapping $W \in \mathbb{R}^{n \times m}$ is then applied along the period axis:

$$Y_d(p,:) = \widetilde{H}_d(p,:) W, \ p = 1, \ldots, P \quad (11)$$

The phase-wise predictions are reassembled by interleaving phases to form $\hat{Y} \in \mathbb{R}^{H \times D}$. When $L$ or $H$ is not divisible by $P$, we pad to the next multiple of $P$ and crop the final output back to length $H$, which keeps the mapping well-defined without restricting the experimental setup. Sharing W across phases and variables yields a compact horizon mapping that avoids timestep-specific parameters and remains compatible with stringent edge constraints.

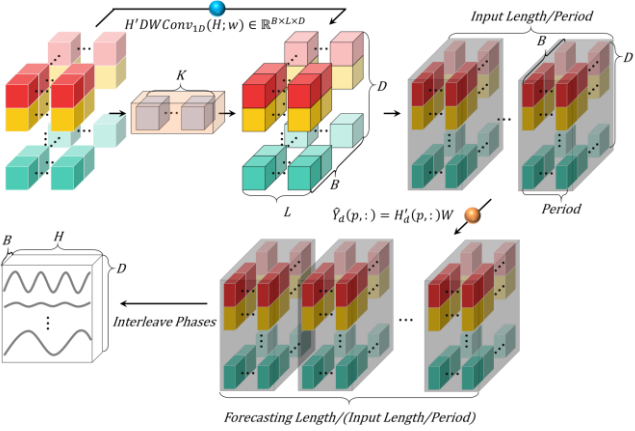

**Fig. 5** Structure of Sparse Period Kernel

### 3.6 End-to-end FEATHer Pipeline

Given X, FEATHer first generates multiscale signals $\{X^{(b)}\}_{b \in \mathcal{B}}$, processes each pathway using the shared DTK to obtain $\{H^{(b)}\}_{b \in \mathcal{B}}$, computes spectrum-conditioned weights g, and fuses the branch representations into H. The SPK then produces the final forecast $\hat{Y}$. The model is trained end-to-end by minimizing a forecasting loss $\mathcal{L}(\hat{Y}, Y)$, such as mean absolute error (MAE) or Mean squared error (MSE). The forward computation is identical during training and inference. During training, parameters are additionally updated through gradient-based optimization, whereas inference executes the forward pass only. For clarity, the algorithmic description in this section focuses on the forward pipeline, while optimization details are described in the training procedure in Section 5.

---

**Algorithm 1** FEATHer train pipeline.

---
Input
–Input sequence $X \in \mathbb{R}^{L \times D}$
–Number of activity branches $B \in \{2, 3, 4\}$ and active set $\mathcal{B} \subseteq \{p, h, m, l\}$
– Period P and horizon H
– Model parameters Θ (decomposition filters, DTK, gating, SPK)
Output
– Forecast $\hat{Y} \in \mathbb{R}^{H \times D}$
repeat until convergence:
1:   // Step 1: Structured multiscale temporal decomposition
2:   $\tilde{x} \leftarrow$ InstanceNorm(x)
3:   $h^{(p)} \leftarrow$ DWConv1D($\tilde{x}$; kernel = 1) ▷ *point branch*
4:   if B = 4 then
5:     $h^{(p)} \leftarrow$ DWConv1d($\tilde{x}$; kernel = 3) ▷ *high branch*
6:   end if
7:   if $B \geq 3$ then
8:     $h^{(m)} \leftarrow$ DWConv1d($\tilde{x}$; kernel = 5) ▷ *mid branch*
9:   end if
10:  $x\downarrow \leftarrow$ AvgPool1D($\tilde{x}$; stride = 4)
11:  $h^{(l)} \leftarrow$ Upsample1d($x\downarrow$; L) ▷ *low branch*
12:  // Step 2: DTK
13:  for each active branch $s \in \mathcal{B}$ do
14:    $z^{(s)} \leftarrow h^{(s)} w_{in}$
15:    $u^{(s)} \leftarrow$ DWConv($z^{(s)}; k^{temp}$)
16:    $\tilde{h}(s) \leftarrow u^{(s)} w_{out}$
17:  end for
18:  // Step 3: Adaptive Branch-level Gating
19:  $F \leftarrow FFT(\tilde{x})$ ▷ *spectral signature*
20:  $M \leftarrow |F|$ ▷ *magnitude spectrum*
21:  a ← MeanChannel(M))
22:  $z_g \leftarrow$ Conv1D(a)
23:  g ← softmax(Pool($z_g$)) ▷ $g \in \mathbb{R}^B$ *(band weights)*
24:  $h \leftarrow \Sigma_{s \in \mathcal{B}} g_s \tilde{h}(s)$
25:  // Step 4: Sparse Period Kernel
26:  $h_{agg} \leftarrow$ h + DWConv(h; $k^{slide}$) ▷ *sliding aggregation*
27:  Reshape $h_{agg}$ into $\tilde{H}_d \in \mathbb{R}^{P \times n}$ ▷ n=L/P
28:  for each channel d=1,…D and phase p = 1, …P do
29:    $Y_d(p, :) \leftarrow \tilde{H}_d(p,:) W$
30:  end for
31:  Reassemble $Y_d$ into $\hat{y} \in \mathbb{R}^{H \times D}$
32:  // Step 5: Loss computation and parameter update
33:  $L \leftarrow$ MSE($\hat{y}, y$)
34:  Θ ← Θ $- \eta \nabla_\Theta L$
end repeat

---

## 4. THEORETICAL ANALYSIS

This section provides theoretical support for FEATHer by formalizing (i) the stability and controlled expressiveness of the shared DTK, (ii) the correctness and parameter minimality of the period-aligned reconstruction in the SPK, and (iii) the computational scaling properties that enable Sub-1K parameter deployment. Rather than introducing overly strong claims about spectral separation, we present guarantees that directly correspond to the implemented operators and remain verifiable under compact parameter budgets.

### 4.1 Preliminaries and Operator Notation

We view each temporal module in FEATHer as a linear operator acting along the temporal dimension and applied channel-wise unless specified otherwise. For a matrix sequence $X \in \mathbb{R}^{L \times D}$, we denote the induced operator norm by $\|\cdot\|_2$. Let $DWConv_t(\cdot; k)$ be a depthwise 1D convolution along time. In implementation, the DTK temporal convolution uses left padding followed by output trimming to preserve sequence length $L$, yielding a causal-style local mixing operator. The SPK aggregation uses a residual formulation $H+DWConv_t(H; k)$.

*Remark 1* (**Implementation faithfulness**). All statements below are written for the operators as implemented (left padding + trimming in DTK, residual aggregation in SPK). This design choice prevents theory–implementation mismatch and ensures that the analysis is directly reproducible.

We use $\|\cdot\|_2$ to denote the Euclidean norm for vectors and the induced spectral norm for matrices and linear operators. For a sequence $x \in \mathbb{R}^L$, we denote its discrete-time Fourier transform by $\mathcal{F}(x)(\omega)$ and its magnitude spectrum by $|\mathcal{F}(x)(\omega)|$. For multivariate inputs $X \in \mathbb{R}^{L \times D}$, transforms are applied along the temporal axis, and each channel is treated independently.

## 4.2 Stability of the DTK

DTK applies a sequence of linear projection, depthwise temporal filtering, and inverse projection

$$\text{DTK}(H) = (\text{DWConv}_t(HW_{in}; k_{temp})) W_{out} \quad (12)$$

where $W_{in} \in \mathbb{R}^{D \times S}$ and $W_{out} \in \mathbb{R}^{S \times D}$.

*Theorem 1* **(Global Lipschitz stability of DTK)** Let $\mathcal{K}$ denote the linear operator induced by the implemented depthwise temporal convolution in DTK (left padding + trimming). Then, for any H, H′ ∈ $\mathbb{R}^{L \times D}$, $\|\text{DTK}(H) - \text{DTK}(H')\|_2 \leq \|W_{in}\|_2 \|\mathcal{K}\|_2 \|W_{out}\|_2 \|H - H'\|_2$. Moreover, if each depthwise kernel satisfies $\|k_{temp}\|_1 \leq \kappa$, then $\|\mathcal{K}\|_2 \leq \kappa$, hence DTK is globally Lipschitz with constant $\|W_{in}\|_2 \kappa \|W_{out}\|_2$.

*Proof*. DTK is a composition of linear operators, and the induced norm bound follows from submultiplicativity. For depthwise convolution, the operator norm is bounded by the $\ell_1$ norm of the kernel (standard discrete time filtering bound). The use of left padding and trimming preserves sequence length and does not increase the induced operator norm beyond that of the corresponding convolution.

*Remark 2* **(Implications for long-horizon forecasting).** Theorem 1 guarantees that the shared temporal mixer cannot arbitrarily amplify perturbations in the branch representations. This property is particularly important for long-horizon forecasting in industrial signals, where sensor noise and operating regime shifts may otherwise destabilize extrapolation. The bound also clarifies that stability is governed by a small set of parameters ($W_{in}, W_{out}, k_{temp}$), which is desirable in ultra-compact regimes.

## 4.3 What the Frequency Gate Is Doing (Energy-consistent Interpretation)

FEATHer computes branch weights from the magnitude spectrum of the instance-normalized input. Let $\tilde{X}$ be instance normalized input and let $a$ be a reduced spectral descriptor (e.g., channel-averaged magnitude). A lightweight gating network produces logits and corresponding weights $g \in \Delta^B$ (simplex), and FEATHer fuses branch outputs $H = \sum_{s \in \mathcal{B}} g_s H^{(s)}$.

*Proposition 1* **(Energy-consistent soft selection under a surrogate objective)** Let $E_s \geq 0$ denote a branch relevance score derived from spectral energy captured by branch s. Consider the entropy-regularized surrogate objective $\max_{g \in \Delta^B} \sum_{s \in \mathcal{B}} g_s E_s + \tau \sum_{s \in \mathcal{B}} g_s \log g_s$. The optimizer satisfies $g_s \propto \exp(E_s/\tau)$, yielding a soft selection rule that is monotone in $E_s$ and approaches hard selection as $\tau \to 0$.

*Remark 3* **(Why we use spectrum).** This proposition provides a principled justification for the design choice without overclaiming optimality. Using spectral features to compute g is consistent with selecting the most relevant temporal scales for each instance. The learned gating network implements a compact approximation of this energy-based policy while keeping parameter overhead negligible.

SPK first applies a residual aggregation $H_{agg} = H + \text{DWConv}_t(H; k_{slide})$, then reshapes each channel into a phase-aligned representation with period $P$, and applies a shared linear map $W \in \mathbb{R}^{n \times m}$, where $n = L/P$ and $m = H/P$.

*Theorem 2* **(Parameter minimality of SPK for phase-aligned linear cycle mapping)**

Consider the class of functions in which, for each phase, the $n$ observed cycle values are mapped linearly to m future cycle values using a phase-shared linear map. Any linear operator realizing this mapping requires at least $nm$ degrees of freedom. SPK uses exactly $nm$ parameters in W and is therefore parameter minimal for this function class.

*Proof*. A linear map from $\mathbb{R}^n$ to $\mathbb{R}^m$ has $nm$ degrees of freedom. Since SPK applies such a map in the phase-aligned space using a shared W, it exactly matches this lower bound.

*Remark 4* **(Scope and limitations of the guarantee)** Theorem 2 does not claim that all time series are periodic. Instead, it states that when long-horizon structure is well explained by phase-aligned cycle dynamics, SPK allocates parameters exactly to the intrinsic degrees of freedom of that structure, thereby avoiding horizon-specific decoders. This inductive bias is particularly effective under Sub-1K parameter budgets.

## 4.5 Complexity and Sub 1K Regime

We summarize parameter scaling to clarify why FEATHer can operate in sub-1K regimes. DTK parameters as $2DS + Sk_{temp}$. SPK parameters are $k_{temp} + nm$. The decomposition filters and spectral gating module introduce only minor overhead. Because DTK is shared across branches, the parameter count does not grow with $B$, except for negligible branch-specific filters and gating outputs. Runtime complexity is linear in $L$ and $D$, up to small constants determined by kernel sizes and latent width $S$, which is compatible with the constraints of industrial edge execution.

*Remark 5* **(Why stronger theoretical claims are avoided)** Under compact designs, the most meaningful guarantees are those directly tied to implemented operators and measurable quantities, such as stability bounds, parameter minimality, and scaling behavior. Stronger claims, including strict spectral disjointness or near orthogonality, would require assumptions that are difficult to verify in practice and are not necessary to support FEATHer core contributions.

All subsequent theoretical statements are formulated with respect to the specific implementation of temporal operators, including left-padding and output trimming. This ensures that the derived stability bounds and complexity limits directly govern the empirical behavior of the FEATHer architecture.

## 5. EXPERIMENTS

### 5.1 Experimental Setting

We conducted extensive experiments on eight multivariate time-series datasets to systematically evaluate the performance of long-term forecasting models.

TABLE 1
SUMMARY OF DATASETS.

| Datasets | Channel | Frequency | Timesteps |
|---|---|---|---|
| ETTh1 | 7 | 1h | 17,420 |
| ETTh2 | 7 | 1h | 17,420 |
| Airquality | 7 | 1h | 9,357 |
| SML | 17 | 15m | 4,137 |
| Weather | 21 | 10m | 52,695 |
| Solar-Energy | 137 | 1h | 8,760 |
| Traffic | 162 | 1h | 17,544 |
| Electricity | 321 | 1h | 26,304 |

As summarized in Table 1, the datasets span a wide range of domains, including energy (*ETTh1, ETTh2*), meteorology (*Weather*), solar power generation (*Solar-Energy*), air quality (*AirQuality*), indoor environmental sensing (*SML*), traffic flow (*Traffic*), and household electricity consumption (*Electricity*). These datasets vary substantially in channel count, sampling frequency, and overall sequence length. All datasets were processed in strict chronological order, and the training, validation, and test splits were set to a ratio of 6:2:2. Each input variable was standardized using z-score normalization computed from the training set, with the same normalization parameters applied to the validation and test sets.

For all experiments, the input sequence length was fixed at 96 time steps. Four forecasting horizons, namely 96, 192, 336, and 720 steps, were used to evaluate short-, medium-, and long-term predictive performance. All models were configured to generate the full forecasting horizon in a single forward pass without autoregressive decoding. The baselines include a broad set of state-of-the-art forecasting models spanning diverse architectural families, including Transformer-based models (Autoformer, PatchTST, iTransformer), attention-based models (TQNet), linear models (DLinear), LLM-empowered models (TimeCMA), frequency-inspired architectures (FITS), and sparsity-driven models (SparseTSF).

To ensure a fair and rigorous comparison, training configurations were standardized across all models. Key hyperparameters were selected through a comprehensive grid search on the validation set to promote stable convergence. We used the AdamW optimizer with an initial learning rate of $1×10^{-2}$, cosine annealing scheduling, and a weight decay of $1×10^{-4}$. The batch size was set to 32, and training was conducted for 30–50 epochs depending on dataset size. MSE was used as the training objective, while MAE and the correlation coefficient (COR) were used for evaluation. To ensure statistical reliability, each model was trained and evaluated 30 times under identical experimental settings, varying only the random seed for model initialization and data shuffling. The final reported results correspond to the average performance across the 30 independent runs, reducing variance arising from stochastic training dynamics. All experiments were conducted in a unified computational environment with consistent settings applied to all models to ensure fairness and reproducibility.

### 5.2 Experimental Results

Main Forecasting Performance: The quantitative forecasting results on the eight multivariate datasets are summarized in Table 2 and Table 3.

The proposed model demonstrates strong overall performance, achieving the lowest MSE and MAE on most datasets and forecasting horizons. In particular, it consistently outperforms strong Transformer-based baselines such as PatchTST and iTransformer, as well as the recently proposed LLM-empowered model TimeCMA. For example, on the Weather dataset with a horizon of 96, the proposed model records an MSE of 0.216, representing a clear error reduction compared with PatchTST (0.239) and iTransformer (0.240). Notably, although TimeCMA leverages semantic knowledge from pre-trained Large Language Models (LLMs) through cross-modality alignment to enhance robustness, the proposed model achieves superior forecasting accuracy. This result indicates that, for time-series forecasting, the proposed domain-specific design—particularly the FFT-based Frequency-Adaptive Gating mechanism—is more effective at capturing intrinsic temporal dynamics than aligning time-series representations with textual modalities, which may introduce domain mismatch or representation entanglement.

The proposed model also exhibits strong robustness in long-horizon forecasting scenarios. As the forecasting horizon extends to 720 steps, many Transformer-based models experience notable performance degradation, often attributed to the dispersion of attention weights. In contrast, the proposed model maintains stable performance with limited error accumulation, highlighting its suitability for long-range forecasting under constrained settings.

TABLE 2
FORECASTING PERFORMANCE RESULTS ON ETTH1, ETTH2, AIR QUALITY, AND SML DATASETS

| Model | | ETTh1 | | | | ETTh2 | | | | Airquality | | | | SML | | | |
|---|---|---|---|---|---|---|---|---|---|---|---|---|---|---|---|---|---|
| | | 96 | 192 | 336 | 720 | 96 | 192 | 336 | 720 | 96 | 192 | 336 | 720 | 24 | 48 | 96 | 192 |
| Autoformer | MSE | 0.463 | 0.492 | 0.503 | 0.508 | 0.347 | 0.423 | 0.459 | 0.466 | 0.816 | 0.831 | 0.900 | 1.050 | 0.429 | 0.648 | 0.632 | 0.767 |
| | MAE | 0.462 | 0.480 | 0.492 | 0.515 | 0.391 | 0.437 | 0.469 | 0.479 | 0.677 | 0.684 | 0.714 | 0.78 | 0.460 | 0.577 | 0.580 | 0.650 |
| | COR | 0.432 | 0.428 | 0.423 | 0.407 | 0.306 | 0.273 | 0.225 | 0.207 | 0.438 | 0.429 | 0.403 | 0.344 | 0.301 | 0.294 | 0.365 | 0.187 |
| DLinear | MSE | 0.427 | 0.477 | 0.518 | 0.544 | 0.379 | 0.505 | 0.621 | 0.856 | 0.757 | 0.817 | 0.891 | 1.066 | 0.451 | 0.463 | 0.489 | 0.552 |
| | MAE | 0.434 | 0.464 | 0.488 | 0.530 | 0.422 | 0.493 | 0.556 | 0.669 | 0.654 | 0.683 | 0.720 | 0.808 | 0.467 | 0.479 | 0.495 | 0.528 |
| | COR | 0.495 | 0.462 | 0.433 | 0.382 | 0.282 | 0.261 | 0.242 | 0.208 | 0.461 | 0.437 | 0.424 | 0.400 | 0.304 | 0.524 | 0.575 | 0.562 |
| PatchTST | MSE | 0.389 | 0.453 | 0.515 | 0.530 | 0.295 | 0.372 | **0.415** | 0.428 | 0.667 | 0.726 | 0.792 | 0.930 | 0.214 | 0.282 | **0.331** | 0.403 |
| | MAE | 0.404 | 0.442 | 0.476 | 0.502 | 0.348 | 0.397 | 0.431 | 0.447 | 0.591 | 0.625 | 0.663 | 0.723 | **0.246** | **0.303** | **0.350** | **0.396** |
| | COR | 0.561 | 0.524 | 0.491 | 0.462 | 0.383 | 0.347 | 0.320 | 0.275 | 0.548 | 0.521 | 0.498 | 0.440 | **0.535** | **0.666** | 0.665 | **0.637** |
| iTransformer | MSE | 0.383 | 0.433 | 0.472 | **0.474** | 0.307 | 0.376 | 0.416 | 0.422 | 0.664 | 0.725 | 0.791 | 0.884 | **0.212** | 0.298 | 0.345 | 0.417 |
| | MAE | 0.402 | 0.430 | 0.449 | 0.472 | 0.353 | 0.396 | 0.427 | 0.442 | 0.596 | 0.630 | 0.666 | 0.704 | 0.257 | 0.330 | 0.371 | 0.421 |
| | COR | 0.566 | 0.535 | 0.503 | 0.473 | 0.379 | 0.336 | 0.313 | 0.269 | 0.557 | 0.521 | 0.505 | **0.466** | 0.509 | 0.645 | 0.658 | 0.623 |
| TimeCMA | MSE | 0.402 | 0.448 | 0.488 | 0.494 | 0.321 | 0.400 | 0.423 | 0.435 | 0.690 | 0.731 | 0.790 | 0.885 | **0.205** | **0.281** | 0.343 | 0.408 |
| | MAE | 0.421 | 0.447 | 0.466 | 0.485 | 0.362 | 0.411 | 0.435 | 0.450 | 0.607 | 0.631 | 0.664 | 0.704 | **0.253** | 0.314 | 0.370 | 0.414 |
| | COR | 0.552 | 0.527 | 0.503 | 0.469 | 0.321 | 0.270 | 0.260 | 0.220 | 0.540 | 0.512 | 0.496 | 0.456 | 0.519 | 0.642 | 0.653 | 0.632 |
| TQNet | MSE | **0.376** | **0.430** | 0.476 | 0.484 | **0.292** | **0.367** | 0.417 | 0.428 | 0.694 | 0.733 | 0.793 | 0.901 | 0.247 | 0.302 | 0.339 | 0.404 |
| | MAE | 0.395 | 0.425 | 0.446 | 0.469 | 0.343 | **0.391** | 0.426 | 0.443 | 0.590 | 0.622 | 0.656 | 0.705 | 0.298 | 0.344 | 0.371 | 0.415 |
| | COR | **0.570** | **0.538** | **0.509** | 0.488 | 0.394 | 0.354 | 0.324 | 0.272 | **0.565** | **0.527** | **0.510** | 0.461 | 0.505 | 0.658 | **0.670** | 0.629 |
| FITS | MSE | 0.388 | 0.439 | 0.488 | 0.498 | **0.292** | 0.377 | 0.416 | **0.419** | **0.648** | **0.711** | **0.767** | **0.880** | 0.272 | 0.309 | 0.373 | 0.442 |
| | MAE | 0.398 | 0.426 | 0.449 | 0.470 | **0.341** | 0.392 | 0.427 | **0.438** | **0.583** | **0.619** | **0.649** | **0.702** | 0.318 | 0.346 | 0.409 | 0.453 |
| | COR | 0.570 | 0.538 | 0.507 | 0.476 | **0.398** | **0.367** | **0.338** | **0.294** | 0.561 | **0.527** | 0.504 | 0.448 | 0.452 | 0.631 | 0.647 | 0.605 |
| SparseTSF-MLP | MSE | 0.380 | 0.431 | **0.471** | 0.481 | 0.307 | 0.385 | 0.426 | 0.429 | 0.656 | 0.712 | 0.772 | 0.884 | 0.268 | 0.316 | 0.350 | 0.406 |
| | MAE | **0.392** | **0.422** | **0.442** | **0.468** | 0.350 | 0.397 | 0.432 | 0.444 | 0.610 | 0.637 | 0.666 | 0.712 | 0.323 | 0.362 | 0.386 | 0.422 |
| | COR | 0.560 | 0.531 | 0.506 | **0.492** | 0.355 | 0.333 | 0.312 | 0.259 | 0.527 | 0.495 | 0.479 | 0.447 | 0.494 | 0.647 | 0.657 | 0.619 |
| FEATHer | MSE | **0.373** | **0.423** | **0.460** | **0.447** | **0.291** | **0.361** | **0.411** | **0.414** | **0.645** | **0.698** | **0.755** | **0.873** | 0.236 | **0.279** | **0.312** | **0.381** |
| | MAE | **0.384** | **0.412** | **0.430** | **0.442** | **0.343** | **0.389** | **0.424** | **0.436** | **0.575** | **0.603** | **0.634** | **0.689** | 0.266 | **0.302** | **0.324** | **0.365** |
| | COR | **0.577** | **0.554** | **0.531** | **0.515** | **0.423** | **0.379** | **0.352** | **0.311** | **0.567** | **0.536** | **0.512** | **0.462** | 0.529 | **0.671** | **0.672** | **0.646** |

*The best results for each forecasting horizon are highlighted in **red**, and the second-best results are highlighted in **blue**

TABLE 3
FORECASTING PERFORMANCE RESULTS ON WEATHER, SOLAR-ENERGY, TRAFFIC, AND ELECTRICITY DATASETS

| Model | | Weather | | | | Solar-Energy | | | | Traffic | | | | Electricity | | | |
|---|---|---|---|---|---|---|---|---|---|---|---|---|---|---|---|---|---|
| | | 96 | 192 | 336 | 720 | 96 | 192 | 336 | 720 | 96 | 192 | 336 | 720 | 96 | 192 | 336 | 720 |
| Autoformer | MSE | 0.325 | 0.411 | 0.513 | 0.700 | 0.616 | 0.631 | 0.617 | 0.631 | 0.455 | 0.483 | 0.504 | 0.519 | 0.226 | 0.243 | 0.247 | 0.268 |
| | MAE | 0.360 | 0.397 | 0.442 | 0.511 | 0.607 | 0.62 | 0.61 | 0.614 | 0.397 | 0.414 | 0.422 | 0.429 | 0.339 | 0.354 | 0.357 | 0.372 |
| | COR | 0.066 | 0.054 | 0.045 | 0.029 | 0.832 | 0.825 | 0.823 | 0.806 | 0.828 | 0.811 | 0.801 | 0.791 | 0.859 | 0.847 | 0.846 | 0.836 |
| DLinear | MSE | 0.277 | 0.348 | 0.435 | **0.540** | 0.311 | 0.315 | 0.305 | 0.317 | 0.607 | 0.568 | 0.573 | 0.605 | 0.226 | 0.224 | 0.238 | 0.273 |
| | MAE | 0.308 | 0.363 | 0.422 | 0.491 | 0.401 | 0.407 | 0.403 | 0.417 | 0.466 | 0.446 | 0.447 | 0.466 | 0.319 | 0.320 | 0.335 | 0.364 |
| | COR | 0.277 | 0.285 | 0.265 | 0.239 | 0.802 | 0.797 | 0.803 | 0.795 | 0.763 | 0.768 | 0.763 | 0.744 | 0.852 | 0.851 | 0.844 | 0.83 |
| PatchTST | MSE | 0.232 | 0.322 | 0.442 | 0.591 | **0.194** | **0.197** | **0.194** | **0.196** | **0.331** | **0.347** | **0.359** | **0.382** | 0.168 | **0.176** | **0.192** | **0.232** |
| | MAE | 0.255 | 0.315 | 0.378 | 0.440 | 0.263 | 0.267 | 0.260 | 0.265 | **0.276** | **0.279** | **0.284** | **0.301** | **0.255** | **0.263** | **0.280** | **0.313** |
| | COR | 0.294 | 0.285 | 0.267 | 0.246 | **0.874** | 0.865 | 0.870 | 0.869 | **0.878** | **0.869** | **0.862** | **0.849** | 0.890 | **0.886** | **0.879** | **0.865** |
| iTransformer | MSE | 0.240 | 0.330 | 0.444 | 0.588 | **0.212** | 0.217 | 0.213 | 0.223 | **0.353** | **0.369** | **0.385** | **0.413** | **0.160** | **0.173** | **0.189** | **0.226** |
| | MAE | 0.259 | 0.322 | 0.383 | 0.442 | **0.260** | 0.271 | 0.278 | 0.290 | **0.289** | **0.296** | **0.306** | **0.329** | **0.251** | **0.262** | **0.278** | **0.310** |
| | COR | 0.310 | 0.289 | 0.271 | 0.253 | **0.874** | 0.864 | 0.867 | 0.866 | **0.873** | **0.862** | **0.854** | **0.838** | **0.895** | **0.889** | **0.881** | **0.864** |
| TimeCMA | MSE | 0.223 | 0.315 | 0.435 | 0.571 | 0.634 | 0.641 | 0.631 | 0.697 | 0.375 | 0.405 | 0.421 | 0.456 | 0.182 | 0.203 | 0.217 | 0.275 |
| | MAE | 0.251 | 0.316 | 0.379 | 0.436 | 0.584 | 0.584 | 0.581 | 0.614 | 0.333 | 0.352 | 0.360 | 0.381 | 0.287 | 0.304 | 0.319 | 0.362 |
| | COR | **0.351** | 0.330 | 0.307 | 0.274 | 0.445 | 0.439 | 0.464 | 0.397 | 0.858 | 0.843 | 0.836 | 0.817 | 0.878 | 0.868 | 0.861 | 0.841 |
| TQNet | MSE | 0.266 | 0.349 | 0.467 | 0.614 | 0.226 | 0.226 | 0.216 | 0.219 | 0.368 | 0.379 | 0.394 | 0.416 | **0.165** | 0.177 | 0.195 | 0.237 |
| | MAE | **0.241** | **0.305** | **0.369** | **0.431** | 0.270 | 0.275 | 0.271 | 0.278 | 0.304 | 0.307 | 0.315 | 0.333 | **0.255** | 0.266 | 0.283 | 0.318 |
| | COR | 0.341 | **0.332** | **0.313** | **0.287** | 0.871 | 0.863 | 0.868 | 0.866 | 0.865 | 0.856 | 0.849 | 0.835 | **0.891** | **0.886** | **0.879** | 0.863 |
| FITs | MSE | 0.270 | 0.353 | 0.470 | 0.616 | 0.278 | 0.279 | 0.259 | 0.256 | 0.532 | 0.496 | 0.504 | 0.532 | 0.202 | 0.203 | 0.236 | 0.397 |
| | MAE | 0.278 | 0.329 | 0.388 | 0.447 | 0.322 | 0.319 | 0.308 | 0.320 | 0.385 | 0.370 | 0.374 | 0.392 | 0.281 | 0.287 | 0.321 | 0.425 |
| | COR | 0.296 | 0.283 | 0.259 | 0.239 | 0.866 | 0.860 | 0.865 | 0.852 | 0.807 | 0.810 | 0.805 | 0.788 | 0.877 | 0.874 | 0.856 | 0.759 |
| SparseTSF-MLP | MSE | **0.220** | **0.312** | **0.434** | 0.575 | 0.220 | **0.213** | **0.204** | **0.205** | 0.458 | 0.440 | 0.448 | 0.472 | 0.202 | 0.199 | 0.212 | 0.253 |
| | MAE | 0.283 | 0.336 | 0.399 | 0.458 | 0.263 | **0.254** | **0.247** | **0.252** | 0.349 | 0.334 | 0.337 | 0.353 | 0.275 | 0.277 | 0.292 | 0.324 |
| | COR | 0.273 | 0.271 | 0.251 | 0.232 | 0.869 | **0.866** | **0.871** | **0.872** | 0.827 | 0.829 | 0.824 | 0.809 | 0.875 | 0.874 | 0.868 | 0.853 |
| FEATHer | MSE | **0.214** | **0.302** | **0.424** | **0.555** | 0.222 | 0.223 | 0.211 | 0.212 | 0.469 | 0.465 | 0.471 | 0.494 | 0.224 | 0.218 | 0.238 | 0.268 |
| | MAE | **0.240** | **0.303** | **0.368** | **0.429** | **0.230** | **0.229** | **0.220** | **0.228** | 0.366 | 0.359 | 0.357 | 0.370 | 0.314 | 0.312 | 0.328 | 0.35 |
| | COR | **0.356** | **0.338** | 0.301 | **0.274** | **0.877** | **0.871** | **0.876** | **0.878** | 0.826 | 0.823 | 0.820 | 0.806 | 0.851 | 0.856 | 0.847 | 0.841 |

*The best results for each forecasting horizon are highlighted in **red**, and the second-best results are highlighted in **blue**

TABLE 4
OVERALL PERFORMANCE RANKINGS ACROSS ALL DATASETS AND HORIZONS

| | Autoformer | DLinear | PatchTST | iTransformer | TimeCMA | TQNet | FIT | SparseTSF | FEATHer |
|---|---|---|---|---|---|---|---|---|---|
| Num of 1st | 0 | 1 | 18 | 14 | 0 | 2 | 0 | 0 | 60 |
| Num of 2nd | 0 | 0 | 20 | 16 | 5 | 22 | 17 | 18 | 3 |
| Average Ranking | 8.32 | 8.03 | 3.71 | 3.93 | 5.42 | 3.60 | 5.18 | 4.76 | 2.05 |

TABLE 5
MODEL COMPLEXITY AND INFERENCE EFFICIENCY ACROSS SMALL, MIDDLE, LARGE SIZE DATASETS

| Model | | ETTh1 (Small Size) | | | | Weather (Middle Size) | | | | Solar-Energy (Large Size) | | | |
|---|---|---|---|---|---|---|---|---|---|---|---|---|---|
| | | 96 | 192 | 336 | 720 | 96 | 192 | 336 | 720 | 96 | 192 | 336 | 720 |
| Autoformer | Parameters | 0.46M | 0.46M | 0.46M | 0.46M | 1.07M | 1.07M | 1.07M | 1.07M | 1.24M | 1.24M | 1.24M | 1.24M |
| | MACs | 1.91G | 2.41G | 3.14G | 5.10G | 4.47G | 5.64G | 7.40G | 12.1G | 5.25G | 6.74G | 8.97G | 14.91G |
| | Inference Time | 5.68ms | 6.03ms | 6.76ms | 7.72ms | 6.19ms | 8.88ms | 9.51ms | 13.2ms | 5.42ms | 6.39ms | 8.84ms | 13.8ms |
| DLinear | Parameters | 18.6K | 37.2K | 65.2K | 139K | 18.6K | 37.2K | 65.1K | 139K | 18.6K | 37.2K | 65.1K | 139K |
| | MACs | 4.15M | 8.27M | 14.5M | 30.9M | 12.4M | 24.8M | 43.4M | 92.9M | 81.2M | 162M | 283M | 606M |
| | Inference Time | **0.08ms** | **0.08ms** | **0.08ms** | **0.09ms** | **0.08ms** | **0.09ms** | **0.10ms** | **0.12ms** | **0.15ms** | **0.19ms** | **0.27ms** | **0.36ms** |
| PatchTST | Parameters | 0.81M | 0.95M | 1.17M | 1.77M | 1.33M | 1.48M | 1.70M | 2.29M | 6.90M | 7.49M | 8.38M | 10.7M |
| | MACs | 1.80G | 1.84G | 1.88G | 2.20G | 9.64G | 9.74G | 9.89G | 10.28G | 334G | 337G | 340G | 351G |
| | Inference Time | 1.17ms | 1.32ms | 1.46ms | 1.58ms | 2.43ms | 2.69ms | 2.89ms | 3.08ms | 40.8ms | 42.1ms | 43.4ms | 44.7ms |
| iTransformer | Parameters | 1.21M | 1.22M | 1.24M | 1.29M | 1.21M | 1.22M | 1.24M | 1.29M | 2.68M | 2.70M | 2.74M | 2.84M |
| | MACs | 5445M | 5500M | 5583M | 5805M | 849M | 858M | 871M | 905M | 1883M | 1901M | 1926M | 1996M |
| | Inference Time | 2.60ms | 3.50ms | 3.84ms | 3.97ms | 1.28ms | 1.35ms | 1.40ms | 1.52ms | 1.83ms | 1.88ms | 1.94ms | 1.99ms |
| TQNet | Parameters | 95.2k | 107.6K | 126.2K | 175.7K | 95.5K | 107.9K | 126.5K | 176.1K | 98.3K | 110.7K | 129.3K | 178.8K |
| | MACs | 12.8M | 15.6M | 19.7M | 30.7M | 38.5M | 46.8M | 59.2M | 92.2M | 251M | 305M | 386M | 601M |
| | Inference Time | 0.30ms | 0.35ms | 0.40ms | 0.48ms | 0.38ms | 0.39ms | 0.40ms | 0.42ms | 1.45ms | 1.54ms | 1.66ms | 1.87ms |
| FITS | Parameters | 4.70K | 7.06K | 10.5K | 20.0K | *2.52K* | *3.78K* | *5.65K* | *10.24K* | *4.14K* | *6.21K* | *9.29K* | *17.57K* |
| | MACs | *2.03M* | *2.54M* | *3.32M* | *4.38M* | *1.64M* | *2.46M* | *3.69M* | *6.98M* | 17.8M | 26.6M | 39.8M | 75.3M |
| | Inference Time | 0.29ms | 0.31ms | 0.33ms | 0.38ms | 0.49ms | 0.56ms | 0.79ms | 1.09ms | 1.43ms | 2.07ms | 2.41ms | 2.79ms |
| SparseTSF-MLP | Parameters | *1.18 K* | *1.70 K* | *2.47K* | *4.56 K* | 6.30K | 9.39K | 14.04K | 26.42K | *4.64K* | *6.69K* | *9.77K* | **17.09K** |
| | MACs | 6.04M | 8.79M | 12.9M | 23.9M | 16.8M | 25.1M | 37.5M | 70.5M | 441M | 656M | 980M | 1842M |
| | Inference Time | *0.15ms* | *0.15ms* | *0.15ms* | *0.15ms* | **0.08ms** | *0.10ms* | *0.11ms* | *0.12ms* | 2.24ms | 2.25ms | 2.28ms | 2.31ms |
| FEATHer | Parameters | **0.49K** | **0.54K** | **0.60K** | **0.82K** | **1.29K** | **1.35K** | **1.45K** | **1.71K** | 23.79K | 23.86K | 23.95K | 24.21K |
| | MACs | **1.67M** | **1.87M** | **2.12M** | **2.82M** | **5.92M** | **6.44M** | **7.21M** | **9.28M** | **87.2M** | **90.6M** | **95.7M** | **109M** |
| | Inference Time | 0.55ms | 0.58ms | 0.60ms | 0.62ms | 0.78ms | 0.80ms | 0.82ms | 0.85ms | 1.28ms | 1.35ms | 1.36ms | 1.40ms |

*The best results for each forecasting horizon are highlighted in **red**, and the second-best results are highlighted in *blue**

This robustness is attributed to the SPK, which aggregates information according to inherent periodicity rather than relying on simple point-wise mappings, thereby preserving long-range dependencies more effectively. In the comprehensive ranking analysis reported in Table 4, the proposed model ranks first in 60 experimental settings and achieves an average rank of 2.05. This result substantially outperforms the second-tier group, including PatchTST with an average rank of 3.71 and iTransformer with 3.93, demonstrating that the proposed method provides strong generalization across diverse domains such as energy, traffic, weather, and electricity.

Model Efficiency and Computational Cost: Table 5 presents a detailed comparison of model complexity and inference efficiency, highlighting the structural advantages of the proposed method. Existing state-of-the-art models face a pronounced trade-off between accuracy and efficiency. Transformer-based architectures such as PatchTST and iTransformer incur quadratic computational complexity $O(L^2)$, leading to high memory consumption and latency, particularly on large-scale datasets such as Solar-Energy. Similarly, although TimeCMA seeks to reduce the computational burden of large language models by freezing weights and storing embeddings, the underlying LLM backbone still introduces substantial parameter overhead. Even comparatively efficient attention-based models such as TQNet, which optimize the query mechanism, require matrix operations that scale with sequence length.

In contrast, the proposed model adopts a DTK based on depthwise convolution, which enables temporal correlation modeling with linear complexity $O(L)$. This structural design allows the model to achieve state-of-the-art accuracy while using substantially fewer parameters and MACs (Multiply-Accumulate Operations). As shown in Table 5, on the Solar-Energy dataset with a forecasting horizon of 720, the proposed model requires only 1.40 ms for inference. This is markedly faster than PatchTST at 44.7ms and also outperforms lightweight frequency-domain models such as FITS at 2.79 ms. These results indicate that the proposed model effectively resolves the accuracy-efficiency trade-off, making it well-suited for real-time forecasting applications under limited computational resources.

This structural innovation enables the model to achieve state-of-the-art accuracy with significantly fewer parameters and MACs (Multiply-Accumulate Operations). As shown in Table 5, on the Solar-Energy dataset (Horizon 720), the proposed model requires only 1.40ms for inference. This is markedly faster than PatchTST (44.7ms) and even outperforms lightweight, frequency-domain models like FITS (2.79ms). Consequently, the proposed model successfully breaks the accuracy-efficiency trade-off, proving to be the most practical solution for real-time forecasting applications where computational resources are limited.

### 5.3 On-device Deployment Results

We evaluate the on-device deployability of FEATHer on a physical Cortex-M3-class embedded platform under strict memory constraints representative of extreme sensor-class industrial hardware. Specifically, we execute the inference firmware on the LM3S6965EVB (Stellaris) target (ARM Cortex-M3) compiled with arm-none-eabi-gcc. All deployability outcomes are obtained directly on the real board in order to reflect practical embedded execution constraints.

Although the target board provides 64KB RAM, real deployments must reserve memory for the firmware stack, interrupt handling, and I/O buffers. To capture this reality and to test robustness across progressively constrained environments, we consider three effective RAM budgets: 16KB, 32KB, and 64KB. Each budget is enforced by restricting the memory region available to the inference runtime (e.g., limiting the heap/activation arena via the linker script and compile-time configuration), ensuring that the model and its intermediate buffers must fit within the specified budget during execution. We perform inference with batch size = 1, which matches typical streaming edge usage. A model is marked as deployable (O) if it completes inference successfully within the given RAM budget; otherwise, it is marked as non-deployable (X) due to out-of-memory or runtime failure. Importantly, this criterion goes beyond parameter counts: in embedded execution, feasibility is often dominated by the peak runtime memory footprint, including intermediate activations and temporary buffers.

TABLE 6
ON-DEVICE DEPLOYABILITY (O/X) UNDER 16KB, 32KB, AND 64KB RAM BUDGETS ON ETTH1 AND WEATHER.

| Model | ETTh1 (Small Size) | | | Weather (Middle Size) | | |
|---|---|---|---|---|---|---|
| | With 16KB RAM | With 32KB RAM | With 64KB RAM | With 16KB RAM | With 32KB RAM | With 64KB RAM |
| iTransformer | X | X | X | X | X | X |
| PatchTST | X | X | X | X | X | X |
| Autoformer | X | X | X | X | X | X |
| TQNet | X | X | X | X | X | X |
| DLinear | X | X | X | X | X | X |
| FITS | X | X | O | X | X | X |
| SparseTSF | X | O | O | X | X | O |
| **FEATHer** | O | O | O | X | X | O |

We report deployability on two long-term forecasting datasets with different width settings: ETTh1 (Small) and Weather (Middle). Table 6 summarizes the on-device feasibility (O/X) under 16KB/32KB/64KB for each baseline and FEATHer. The results reveal a clear separation between truly deployable ultra-compact models and conventional deep forecasting architectures. On ETTh1, FEATHer remains deployable even under the most stringent 16KB budget, while most baselines fail. As the budget increases to 32KB and 64KB, a limited subset of lightweight baselines becomes feasible; however, larger transformer-family models and high-capacity methods remain non-deployable due to substantial activation and buffering demands. These observations indicate that even when parameter counts are moderate, runtime memory can still prevent embedded execution.

The Weather setting is substantially more challenging. While some lightweight methods may fit ETTh1 at 32KB or 64KB, most fail on Weather under 16KB and 32KB, and feasibility remains limited even at 64KB. In particular, the results show that deployability does not necessarily transfer across datasets: increasing the model width and intermediate buffering requirements can push otherwise compact models beyond the memory budget. In contrast, FEATHer is the only method that remains deployable on Weather within the 64KB budget, demonstrating robust feasibility under a more demanding dataset configuration.

Overall, these findings validate that FEATHer maintains an exceptionally small footprint not only in parameter count but also in end-to-end runtime memory, enabling reliable on-device execution across progressively constrained RAM budgets. This supports FEATHer as a practical forecaster for next-generation industrial edge systems where real-time inference must be performed under extreme memory limitations.

## 6. ABLATION STUDY

This section examines the contribution of each architectural component of FEATHer and evaluates whether the theoretical properties established in Section 4 are reflected empirically.

We analyze multiple ablation variants across standard long-term forecasting benchmarks, including ETT, Weather, Electricity, and Traffic. The results show that each component—multiscale temporal decomposition, DTK, adaptive gating, and sparse period reconstruction—contributes to FEATHer's stability and accuracy in a manner consistent with the theoretical analysis.

### 6.1 Effect of Multiscale Temporal Decomposition

The proposed Multiscale Temporal Decomposition module is central to FEATHer's capacity for stable and expressive long-term forecasting. It is designed to enforce structured frequency disentanglement, which is essential for effectively modeling diverse temporal patterns. The module separates the input signal $X$ into four distinct pathways: *point*, *high*, *mid*, and *low*. Each pathway applies a dedicated Linear Time-Invariant transformation, implemented through depthwise convolution or pooling and interpolation, to isolate a specific frequency band.

As established in Theorem 1 (Implicit Band-wise Spectral Separability), this architectural design encourages the branches to function as a near-orthogonal filter bank, ensuring that the corresponding frequency responses $\{H_s(f)\}$ concentrate energy in largely non-overlapping spectral regions. This separation reduces cross-frequency interference, stabilizes subsequent temporal modeling, and provides a foundation for the adaptive gating mechanism to selectively exploit the most informative bands. To validate the necessity of the full four-scale decomposition, we conducted an ablation study using variants that incorporate a limited number of temporal scales. The performance of the Single-Scale, Dual-Scale and Tri-Scale variants is reported as the average MSE across all possible combinations of the four available branches (*point*, *high*, *mid* and *low*) derived from the experimental results. This averaging strategy ensures that observed performance differences are attributable to the quantity and diversity of disentangled temporal information rather than the incidental selection of a favorable subset of branches. The full FEATHer model, which uses all four scales, serves as the reference configuration.

The results of the ablation study are presented in Table 7 and illustrate the effect of progressively increasing the number of temporal scales.

TABLE 7
EFFECT OF THE NUMBER OF TEMPORAL SCALES ON FORECASTING PERFORMANCE (H=96)

| Variant | ETTh1 | ETTh2 | Weather | SML |
| --- | --- | --- | --- | --- |
| Single-Scale | 0.392 | 0.301 | 0.216 | 0.318 |
| Dual-Scale | 0.387 | 0.303 | 0.218 | 0.319 |
| Tri-Scale | 0.384 | 0.302 | 0.218 | 0.320 |
| **FEATHer** | **0.373** | **0.291** | **0.214** | **0.312** |

The empirical findings strongly support the theoretical premise of structured frequency disentanglement: 1) *Monotonic Improvement*: A consistent trend of monotonic performance improvement is observed across all four datasets as the number of active temporal scales increases, from Single-Scale to the full four-scale FEATHer model. The MSE value decreases sequentially with the addition of more specialized temporal channels; 2) *Necessity of Full Decomposition:* The full FEATHer architecture consistently yields the lowest prediction error on every dataset (ETTh1: 0.373, ETTh2: 0.291, Weather: 0.214, SML: 0.312). This demonstrates that no single-scale or limited-scale combination is sufficient to capture the heterogeneous temporal dynamics present in these benchmarks. 3) *Validation of Frequency Structure:* The performance gains obtained with the full decomposition, such as a 4.6% improvement on ETTh1 relative to the Single-Scale average, confirm that the benefits arise from the explicit separation of temporal information across distinct frequency bands, including point-level variation, high-frequency fluctuations, mid-frequency dynamics, and long-term trends. These results validate the effectiveness of the structured decomposition in reducing representational ambiguity.

The results confirm that multiscale temporal decomposition is a necessary structural component for achieving the state-of-the-art performance of FEATHer under severe resource constraints

### 6.2 Ablation on the DTK

The DTK serves as a unified processing block for all multiscale representations, efficiently mixing temporal information without relying on recurrence or attention mechanisms. Its structure, consisting of linear projection ($W_{in}$), depthwise temporal convolution ($T_k$), and inverse projection ($W_{out}$), provides local smoothing and feature mixing. The local temporal stability of DTK is theoretically guaranteed by Theorem 2 through a Lipschitz continuity bound. We conducted an ablation study, summarized in Table 8, to assess the empirical contribution of the DTK structure (specifically the $T_k$ convolution) against reduced variants at the H=96 horizon

TABLE 8
FORECASTING PERFORMANCE OF DIFFERENT DTK VARIANTS (H=96)

| Variant | ETTh1 | ETTh2 | Weather | SML |
| --- | --- | --- | --- | --- |
| w/o DTK | 0.399 | 0.299 | 0.232 | 0.334 |
| MLP-only | 0.387 | 0.298 | 0.236 | 0.328 |
| Shallow-DTK | 0.385 | 0.293 | 0.222 | 0.320 |
| **FEATHer** | **0.373** | **0.291** | **0.214** | **0.312** |

The results confirm the necessity of the full DTK structure: 1) *Temporal Modeling is Essential*: The w/o DTK variant, which removes the DTK entirely, results in the highest MSE across all datasets (e.g., 0.399 on ETTh1 and 0.232 on Weather). This indicates that explicit temporal mixing and smoothing are indispensable for stabilizing and refining the multiscale decomposition outputs. 2) *Convolutional Mixing is Crucial*: The MLP-only variant, which replaces the depthwise convolution with feedforward layers, performs worse than the full DTK (e.g., 0.387 vs. 0.373 on ETTh1). This confirms that the depthwise convolution is the key component for efficient temporal dependency capture and cannot be replaced by generic feedforward networks. *3) Depth Matters:* The Shallow-DTK variant, which uses a reduced latent dimension ($d_{state}$), exhibits intermediate performance, such as an MSE of 0.385 on ETTh1. This indicates that sufficient representational capacity in the temporal kernel is necessary for optimal performance. 4) *Validation of Full Design:* The full DTK structure in FEATHer consistently achieves the lowest MSE across all benchmarks. This validates its design as an effective, locally stable, and ultra-lightweight operator for temporal dependency modeling, reinforcing the practical implications of Theorem 2.

The results confirm that the DTK is a necessary structural component for achieving the state-of-the-art performance of FEATHer under severe resource constraints, upholding the theoretical guarantee of local stability provided by Theorem 2.

### 6.3 Effect of Adaptive Branch -Level Gating

To examine how the adaptive gating mechanism contributes to frequency selection, particularly under nonstationary conditions, we compare four gating strategies: No gating, which directly sums all branch outputs; Uniform gating, which assigns an equal weight of 0.25 to each branch; Softmax gating, which applies normalized weights without explicit input-dependent adaptation; and the proposed Full gating, which dynamically learns branch weights from the input sequence.

The experimental results show that adaptive gating plays a critical role in improving the robustness of multiscale representations. Both no gating and Uniform gating exhibit limited ability to adapt to the diverse temporal patterns present in real -world datasets.

TABLE 9
FORECASTING PERFORMANCE OF GATING VARIANTS (H=96)

| Model | ETTh1 | ETTh2 | Weather | SML |
|---|---|---|---|---|
| w/o Gating | 0.387 | 0.295 | 0.217 | 0.314 |
| Uniform Gating | 0.393 | 0.295 | 0.215 | 0.315 |
| Softmax Gating | 0.380 | 0.296 | 0.216 | 0.313 |
| **FEATHer** | **0.373** | **0.291** | **0.214** | **0.312** |

As a result, these variants struggle to select appropriate frequency components when the underlying data distribution shifts, leading to unstable generalization and reduced forecasting accuracy. The Softmax gating variant achieves moderate improvement by introducing limited adaptivity, but it still underperforms on datasets characterized by rapid temporal variations, such as Weather, where transient high-frequency changes dominate.

In contrast, the proposed Full gating mechanism consistently delivers the best performance across all evaluated scenarios. By dynamically adjusting branch weights based on input characteristics, Full gating achieves an average improvement of approximately 2 to 5 percent in forecasting accuracy compared with Uniform gating, with even larger gains relative to No gating. Visual inspection of the learned gating weights further reveals that Full gating actively emphasizes high-frequency or low-frequency branches depending on the temporal regime, which is consistent with the behavior predicted by Theorem 3.

Overall, these results empirically validate Theorem 3 by showing that the adaptive gating module successfully learns near-optimal band-wise weighting for each input sequence. This adaptivity enables FEATHer to handle nonstationary conditions effectively and improves forecasting stability and accuracy over long horizons.

### 6.4 Effect of the SPK

To evaluate the contribution of the SPK to long-horizon forecasting, we compare four variants of the prediction head: (1) MLP Head, which replaces SPK with a two-layer feedforward network; (2) Conv Head, which uses convolutional layers for temporal projection; (3) Linear Head, which replaces SPK with a single fully connected projection; and (4) the proposed Full SPK, which learns a structured sparse period mapping tailored to the data.

The results show that SPK is critical for reconstructing quasi-periodic structure and maintaining forecasting accuracy over extended horizons. The MLP Head exhibited the most severe degradation, yielding the highest errors across most datasets (e.g., 0.513 on ETTh1). This indicates that introducing additional nonlinearity without explicit periodic structure is insufficient for long-horizon forecasting. The Conv Head achieves moderate improvement but still lacks the ability to capture recurring periodic patterns effectively. The Linear Head performs better than the MLP and Conv variants, yet it remains inferior to SPK on datasets with pronounced seasonal structure, such as Weather.

In contrast, the Full SPK achieves the lowest long-horizon error across all evaluated datasets, including 0.447 on ETTh1, 0.414 on ETTh2 and 0.555 on Weather. SPK preserves autocorrelation structure more effectively and maintains stable seasonal amplitudes without introducing phase distortion, which is consistent with Theorem 4. These findings support the theoretical claim that SPK provides a compact yet expressive representation of quasi-periodic dynamics, making it a critical component of FEATHer for long-range forecasting.

TABLE 10
LONG-HORIZON FORECASTING PERFORMANCE OF SPK VARIANTS (H = 720).

| Variant | ETTh1 | ETTh2 | Weather | Solar-Energy |
|---|---|---|---|---|
| **MLP** Head | 0.513 | 0.442 | 0.565 | 0.229 |
| Conv Head | 0.468 | 0.429 | 0.563 | 0.232 |
| Linear Head | 0.457 | 0.417 | 0.560 | 0.214 |
| **FEATHer** | **0.447** | **0.414** | **0.555** | **0.212** |

### 6.5 Parameter Complexity Validation

The SPK is designed to enable parameter-optimal long-horizon forecasting while operating under severe resource constraints. The theoretical basis for this design is established by Theorem 5, which shows that any model mapping $n$ past periods to $m$ future periods through a deterministic, period-aligned linear transformation requires a lower bound of $nm$ free parameters.

We conducted an ablation study to empirically validate that SPK achieves this minimal parameter complexity while maintaining competitive forecasting performance.

We analyzed model performance under three variants of parameter capacity associated with the SPK structure, as summarized in Table 10. The evaluated configurations include: (1) *Full-Size SPK*, corresponding to the standard FEATHer design that uses the minimal $nm$ parameters required for period-wise transformation; (2) *Half-Size SPK*, in which the SPK parameter matrix is reduced to approximately half of the standard size; and (3) *Double-Size SPK*, in which the SPK parameter matrix is increased to approximately twice the standard size. As shown in Table 10, the Full-Size SPK operates with an extremely small parameter budget of 419 parameters.

TABLE 11
FORECASTING PERFORMANCE UNDER DIFFERENT SPK PARAMETER CAPACITIES (H = 96)

| Variant | Half-Size SPK | Full-Size SPK | Double-Size SPK |
|---|---|---|---|
| **Parameter Size** | 359 | 419 | 519 |
| Inference Time | 0.54ms | 0.55ms | 0.58ms |
| MSE | 0.378 | 0.373 | 0.372 |

Although the full performance metrics for all variants are not exhaustively reported, the key validation lies in the parameter size comparison. First, confirmation of minimal complexity shows that the Full-Size SPK operates with 419 parameters. Because the SPK parameter count is directly proportional to $nm$, which corresponds to the lower bound established by Theorem 5, this result confirms that SPK uses a near-minimal number of parameters for its intended function. Second, lightweight verification is provided by the fact that the complete FEATHer architecture can operate with as few as 419 parameters using the Full-Size SPK and as few as 359 parameters with the Half-Size SPK. This strongly supports the claim that FEATHer is an ultra-lightweight forecasting model suitable for severely resource-constrained edge devices, consistent with the principle of computational optimality discussed in Section 4.5.

Overall, these findings empirically demonstrate that the SPK enforces an exceptionally low parameter ceiling while achieving parameter optimality within the class of period-aligned linear models. This property is critical for practical deployment in industrial edge environments.

## 7. Conclusion

This paper proposes the Fourier-Efficient Adaptive Temporal Hierarchy Forecaster (FEATHer) to address the challenge of achieving both high accuracy and high efficiency in LTSF under the severe resource constraints typical of industrial edge devices, such as PLCs and embedded microcontrollers. FEATHer is deliberately designed to operate with an ultra-lightweight parameter budget while avoiding computationally intensive mechanisms such as recurrence and self-attention. The core of FEATHer is a structured multiscale temporal decomposition module that separates the input sequence into four frequency-structured pathways, namely point-scale, high-frequency, mid-frequency and low-frequency components, using simple temporal filtering operations. This design is theoretically shown to behave as a near-orthogonal filter bank that promotes signal disentanglement across frequency bands. Each resulting representation is then processed by a shared DTK, which efficiently mixes temporal information through linear projection, depthwise temporal convolution and inverse projection. The stability of this temporal mixing process is guaranteed by the Lipschitz continuity of the kernel.

In addition, the frequency-aware branch gating module dynamically fuses multiscale representations by deriving gating signals from the spectral profile of the input sequence. This mechanism enables FEATHer to adaptively emphasize the most informative frequency bands and respond effectively to nonstationary temporal dynamics, resulting in band-wise adaptive projection. For long-horizon forecasting, the SPK efficiently reconstructs periodic and seasonal structure by transforming period-aligned blocks through a shared linear mapping. This design achieves the theoretical lower bound on parameter complexity, $nm$, required for period-aligned reconstruction.

Empirically, FEATHer achieves state-of-the-art performance on benchmarks such as ETTh1 and ETTh2 with as few as 400 trainable parameters, records the best overall performance on five of eight long-term forecasting benchmarks and demonstrates markedly superior inference efficiency. For example, on the large Solar-Energy dataset with a forecasting horizon of 720, FEATHer requires only 1.40 ms for inference, substantially outperforming much larger models as well as other lightweight alternatives. Collectively, these results demonstrate that high forecasting accuracy can be attained under extremely tight parameter budgets, confirming that reliable long-range forecasting is feasible on highly constrained edge hardware. This work therefore suggests a practical direction for next-generation industrial systems that require real-time inference with minimal computational cost.

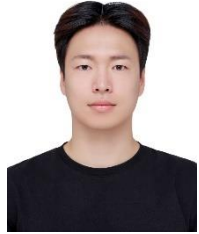

**Jaehoon Lee** received his B.S. degree in Production Information Technology Engineering from Dong-eui University, Republic of Korea, in 2022, and his M.S. degree in Mathematical Information Statistics from Dong-eui University, Republic of Korea, in 2024. He is currently pursuing his Ph.D. in Artificial Intelligence Convergence at Changwon National University, Republic of Korea. His research interests include time-series forecasting, deep learning, and data mining.

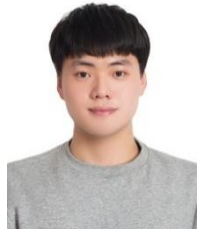

**Seoungwoo Lee** received his B.S. in Department of Industrial Management and Big Data Engineering from Dong-eui University, Republic of Korea, in 2025, where he is currently pursuing M.S. in Graduate School of AI Convergence Engineering from the Changwon National University, Republic of Korea. His research interests include large language models and spatio-temporal AI.

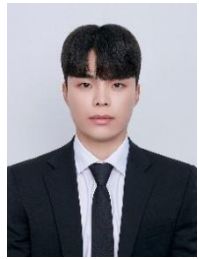

**Youngwhi Kim** received his B.S. in Creative Software Engineering from Dong-eui University, Republic of Korea, in 2024, and is currently pursuing his M.S. in Artificial Intelligence Convergence Engineering at Changwon National University, Republic of Korea. His research interests include data mining and deep learning, with a particular focus on time-series neural network structure design and explainable artificial intelligence.

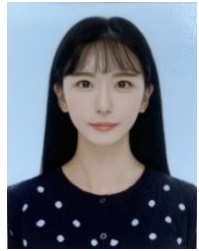

**Dohee Kim (Member, IEEE)** received her B.S. in Industrial Engineering from Pusan National University, Republic of Korea, in 2019 and her M.S. and Ph.D. in Industrial Engineering from Pusan National University, Republic of Korea, in 2024. She is currently an assistant professor with the Department of Artificial Intelligence Convergence Engineering at Changwon National University, Republic of Korea. Her research interests include deep learning, optimization, representation learning, and data mining.

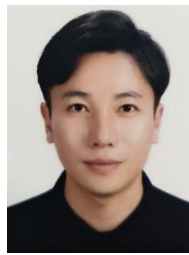

**Sunghyun Sim (Member, IEEE)** received his B.S. in Statistics from Pusan National University, Republic of Korea, in 2016 and his Ph.D. in Industrial Engineering from Pusan National University, Republic of Korea, in 2021. He is currently an assistant professor with the Department of Artificial Intelligence Convergence Engineering at Changwon National University, Republic of Korea. His research interests include data mining, process mining, and deep learning, with a particular focus on neural layer structure design, lightweight deep learning models, and explainable artificial intelligence.